\pdfoutput=1
\documentclass[10pt]{article}
\usepackage[letterpaper]{geometry}
\usepackage{hicss}
\usepackage{times}
\usepackage[T1]{fontenc}
\usepackage[utf8]{inputenc}
\usepackage[none]{hyphenat} 
\usepackage{url}
\usepackage{latexsym}
\usepackage{indentfirst}
\usepackage{graphicx}
\usepackage{booktabs}
\usepackage{newunicodechar}
\graphicspath{{images/}}

\newunicodechar{κ}{\ensuremath{\kappa}}
\newunicodechar{α}{\ensuremath{\alpha}}
\newunicodechar{ρ}{\ensuremath{\rho}}
\newunicodechar{×}{\ensuremath{\times}}
\newunicodechar{−}{\ensuremath{-}}
\newunicodechar{≥}{\ensuremath{\geq}}
\newunicodechar{≤}{\ensuremath{\leq}}
\newunicodechar{≈}{\ensuremath{\approx}}
\newunicodechar{·}{\textperiodcentered}
\newunicodechar{§}{\S}
\newunicodechar{–}{--}
\newunicodechar{—}{---}
\newunicodechar{→}{\ensuremath{\rightarrow}}
\newunicodechar{⁻}{\ensuremath{{}^{-}}}
\newunicodechar{⁸}{\ensuremath{{}^{8}}}
\newunicodechar{“}{``}
\newunicodechar{”}{''}
\newunicodechar{‘}{`}
\newunicodechar{’}{'}

\newcommand{\refitem}[1]{\par\noindent\hangindent=0.3in\hangafter=1 #1\par\vspace{2pt}}

\makeatletter
\newcommand{\acceptancefoot}{\footnotesize This paper has been accepted for the upcoming 60th Hawaii International Conference on System Sciences (HICSS-60)}
\def\ps@empty{\let\@mkboth\@gobbletwo
  \let\@oddhead\@empty \let\@evenhead\@empty
  \def\@oddfoot{\hfil\acceptancefoot\hfil}\let\@evenfoot\@oddfoot}
\makeatother

\title{Whose Voice Survives the Summary? \\ A Voice-Retention Audit of LLM Employee Listening}

\author{Thilo Tamme \\
 Technical University of Munich \\
 {\underline{\vphantom{j}thilo.tamme@tum.de}} \\ \And
 Anton Hantel \\
 Massachusetts Institute of Technology \\
 {\underline{\vphantom{j}hantel@mit.edu}} \\ \And
 Bijan Khosrawi-Rad \\
 Leuphana University Lüneburg \\
 {\underline{\vphantom{j}bijan.khosrawi-rad@leuphana.de}} \\}

\date{}

\begin{document}
\maketitle

\begin{abstract}
Organizations increasingly route employee feedback to leaders through large language model (LLM) summaries, an unaudited layer that silences already-spoken voice. We introduce a Voice Retention / Representation Ratio metric for representational bias in summarization and apply it to a bilingual (English/German) corpus of 2,586 free-text responses from a global professional service company. First, employees supply criticism more reliably than praise (withholding praise is 82 times more common). Second, across 45 leader-summaries the pipeline filters by popularity, not sentiment: criticism survives, yet a concern voiced once is dropped 86\% of the time, with short and German-only content lost on the same axis (theme retention 0.14 vs 0.74; German directional). Controlling for frequency, sentiment has no independent effect; the harm is prevalence-driven, which sentiment-only audits miss. A targeted prompt recovers only named themes. We contribute the metric, field evidence, and a disaggregated voice-retention card.
\end{abstract}

\subsubsection*{Keywords:}

representational harm, summarization bias, responsible AI, employee voice, people analytics

\section{Introduction}

Speaking up at work is risky, and staying silent is costly. Decades of research show that employees routinely withhold concerns from those above them (Morrison \& Milliken, 2000), that voicing problems is interpersonally hazardous (Burris, 2012; Detert \& Edmondson, 2011), and that the resulting silence corrodes organizations (Perlow \& Williams, 2003). Firms have invested heavily in \textit{employee listening} (pulse surveys, sentiment dashboards, always-on feedback) as people analytics (Tursunbayeva et al., 2018). Conversational AI agents are extending this listening even further (Tamme et al., 2026). A new layer now sits atop these channels: large language models (LLMs) that read raw employee comments and summarize them upward for time-constrained leaders. Because instruction-tuned LLM summaries are now judged comparable to human-written ones (Zhang et al., 2024), these pipelines enter production faster than they are scrutinized.

This paper asks: when employee voice is compressed into an LLM summary, which voices are retained and which are systematically dropped? Summarization is not neutral. Abstractive summarizers omit and distort source content (Maynez et al., 2020) and can represent contributors very differently from the source, under-representing demographic groups (Dash et al., 2019) and linguistic varieties (Keswani \& Celis, 2021). When a summary decides which contributions reach a decision-maker, it allocates access, representation, and power; a summary that drops a class of contributors inflicts what Blodgett et al. (2020) call a \textit{representational harm}. We call this failure mode, in employee listening, \textit{silencing-by-summarization}: an algorithmic intermediary that re-suppresses voice after it has cleared the human hurdle of being spoken.

The intuitive worry is that the most valuable voice is the most at risk. Liang et al. (2012) distinguish \textit{promotive} voice (suggestions to improve) from \textit{prohibitive} voice (warnings about harm); the latter carries greater interpersonal risk and tends to arrive hedged and terse (Detert \& Edmondson, 2011), so one would expect a summarizer to smooth away exactly these short, critical comments. Our audit overturns this. Critical voice is, if anything, well retained; what the summary removes is \textit{infrequent} voice, concerns raised in only one or two comments, critical or not. Because a busy leader is unlikely to notice the omission (Parasuraman \& Manzey, 2010), this frequency-based attrition becomes an invisible bias against minority concerns: low-prevalence voice is not a neutral compression artifact but an equity problem about whose concern reaches the decision-maker at all.

The corpus comes from a large global professional service company. Across 2,586 free-text responses, employees answered the IMPROVE prompt 99.8\% of the time but the WENT\_WELL (appreciation) prompt only 80.9\%; withholding praise was roughly 82 times more common than withholding criticism. The classic worry is silence on problems; here the silence falls on the positive channel, so the voice reaching the summarizer is already lopsided toward change.

We pursue two research questions:

\begin{itemize}
\setlength\itemsep{0.2em}
\item \textbf{RQ1.} How is employee voice distributed in naturalistic, bilingual weekly team feedback, specifically the balance between appreciation and change-oriented voice, and between promotive and prohibitive voice?
\item \textbf{RQ2.} When an LLM summarizes this voice upward for leaders, whose voice is retained and whose is dropped (across voice type, language, response length, and theme), and can prompt-level interventions reduce any systematic under-representation?
\end{itemize}

We contribute a reusable \textit{Voice Retention / Representation Ratio} metric for auditing representational bias in abstractive summarization, and field evidence that LLM leader-summaries act as a \textit{popularity filter}, under-representing low-frequency, terse, and non-majority-language voice while retaining criticism; a field operationalization of the promotive/prohibitive distinction and the appreciation asymmetry on naturalistic bilingual data; and prompt-level mitigations with design principles, including a disaggregated voice-retention card, for listening systems that do not silence by summarizing.

\section{Related Work}

\subsection{Employee voice, silence, and psychological safety}

Employee \textit{voice} (discretionary, improvement-oriented communication directed upward) and \textit{silence} (the withholding of such input) are now theorized as distinct constructs, not poles of one scale (Morrison \& Milliken, 2000; Van Dyne et al., 2003; Sherf et al., 2021). Van Dyne et al. (2003) distinguish \textit{acquiescent}, \textit{defensive}, and \textit{prosocial} silence by motive; Morrison and Milliken (2000) show how organizational silence becomes a self-reinforcing, climate-level condition (an organization-wide property, not an individual choice) sustained by managers' aversion to negative feedback. Sherf et al. (2021) find voice and silence empirically independent and silence the stronger correlate of burnout, so staying silent is consequential in itself.

The distinction our study most depends on is Liang et al.'s (2012) split between \textit{promotive} voice (“suggestions for improving the overall functioning” of the unit) and \textit{prohibitive} voice (“concern about work practices, incidents, or employee behavior that are harmful”). Prohibitive voice carries greater interpersonal risk and tends to surface hedged: Detert and Edmondson (2011) document \textit{implicit voice theories} (self-protective rules triggering self-censorship), and Burris (2012) shows managers rate challengers less favorably even while endorsing voice in principle. Psychological safety (Edmondson, 1999; Edmondson \& Bransby, 2023) is the climate gating whether these risks are taken.

\subsection{How voice is transformed as it travels upward}

Most voice research models a direct employee-to-manager exchange; a smaller stream asks what happens as voice is \textit{aggregated and relayed}. Detert et al. (2013) show that how voice flows to and around leaders changes its usefulness, and Burris et al. (2025) isolate \textit{secondhand} accounts (input reaching a decision-maker through an intermediary) and their credibility problem. An LLM that compresses many comments into one summary for leaders is exactly such an intermediary, but an automated one that no prior study has examined.

\subsection{Faithfulness and fairness in LLM summarization}

Abstractive summarizers omit and distort: Maynez et al. (2020) show that neural summarizers hallucinate pervasively, and that entailment-based measures track faithfulness far better than surface-overlap metrics like ROUGE.

The pivot from “is the summary accurate?” to “\textit{whose} contribution is retained?” comes from fairness-in-summarization. Dash et al. (2019) show on real user-generated text that summarizers represent socially salient groups very differently from their source distribution, shifting with summary length; Keswani and Celis (2021) show the analogous effect for \textit{dialects}; Blodgett et al. (2020) name such failures \textit{representational harms}. Joshi et al. (2020) establish \textit{language} as a fairness axis; because German is high-resource, an English/German gap would be notable for arising between two well-resourced languages. None of this work addresses employee voice or the promotive/prohibitive axis.

\subsection{LLM-as-coder and automation bias}

Our method uses LLMs to code constructs at scale, a debated practice: Gilardi et al. (2023) report zero-shot ChatGPT exceeding crowd-worker accuracy at a fraction of the cost, while Ziems et al. (2024) find only \textit{fair} agreement on taxonomic labeling and urge augmentation over replacement. We therefore validate against human adjudication. Automation bias explains why this matters in deployment: people over-trust automated aids and commit errors of omission (Parasuraman \& Manzey, 2010), so a leader is unlikely to notice what a digest dropped.

\subsection{People analytics and governance}

An LLM summarizing weekly employee comments is \textit{people analytics} (Tursunbayeva et al., 2018) and inherits that field's critique: Giermindl et al. (2022) catalog perils an opaque, dissent-dropping summary would aggravate, notably impaired transparency and the marginalization of human reasoning, and Gal et al. (2020) argue people analytics should be a “fallible companion technology.” Bernstein's (2012) transparency paradox and Zuboff's (2015) surveillance-capitalism lens sharpen the stakes of rendering unguarded employee text as managerial data. We model our “voice-retention card” on model cards (Mitchell et al., 2019) and datasheets (Gebru et al., 2021). LLMs are also moving upstream into the collection of employee voice, with conversational agents now conducting qualitative interviews at organizational scale (Tamme et al., 2026), widening the AI-native layer this audit targets.

\subsection{Research gap}

Three gaps converge. Voice/silence theory models silence as a human, climate-level phenomenon and has not examined an \textit{algorithmic} aggregation layer. Summarization-fairness work demonstrates representational harm for demographic groups and dialects on social media, but not for \textit{employee voice} or the \textit{promotive/prohibitive} axis. People-analytics critique remains largely conceptual, lacking both a \textit{measured} harm and a constructive, \textit{auditable} mitigation. Our research questions address these in turn.

\section{Data and Context}

The corpus is the free-text portion of a weekly team-pulse survey run inside a large global professional service company. Most of the survey is Likert items on team satisfaction and project work; its final two questions are optional free-text fields, the only data we analyze. These yield 2,586 free-text responses across 538 project codes, which resolve to 260 distinct teams once the weekly wave suffix is stripped. Each carries the two free-text answers plus its project code, with no demographics, role, seniority, or calendar timestamps. Row order within a project is chronological, so the data form a short panel indexed by wave rank rather than date (95\% of rows carry a recoverable wave number, up to ten waves for the longest projects). The text is bilingual: mostly English, a minority German, a handful mixing the two.

\subsection{The employee-listening instrument}
The two fields are WENT\_WELL (what went well) and IMPROVE (what should change); only about 16\% of participants fill at least one of them (Section~8.4). They map onto Section~2.1's appreciation-versus-change distinction (WENT\_WELL appreciative, IMPROVE promotive/prohibitive; Liang et al., 2012). Because the same person answers both in one sitting, we observe which channel they fill and which they leave blank, the raw material for the Section~5 appreciation asymmetry. A single team produces dozens of comments per wave, which is why a summarization layer is inserted at all.

\subsection{Ethics, anonymization, and provenance}
Our empirical site is described only as a large global professional service company; all client, person, and author identities are removed for review.
The firm that owns this internal feedback authorized its use for this research. Accordingly, the data were anonymized before analysis, are held under controlled access, and no raw employee text is released. We document the corpus following the spirit of datasheets for datasets (Gebru et al., 2021). The surveillance and privacy implications of analyzing unguarded employee text are taken up in Section~8 (Bernstein, 2012; Zuboff, 2015; Giermindl et al., 2022).

\section{Method}

\subsection{The construct framework}
Our unit of analysis is one filled free-text field, a single WENT\_WELL or IMPROVE answer (the 4,673 units are 2,093 WENT\_WELL plus 2,580 IMPROVE). A field raising several points is coded once at its dominant register, so unit counts and the promotive-to-prohibitive ratio are field-level. We code each unit against a 17-key codebook whose spine is the promotive/prohibitive split from Liang et al. (2012): voice type (appreciative, promotive, prohibitive, descriptive-neutral, perfunctory), directness, an explicit-suggestion flag, target (peer, lead, senior manager, client, process, or self), sentiment and intensity, a futility marker (“again,” “still”), and norm-invocation (appeals to a teaming norm such as protected time). A dictionary layer tags fifteen recurring themes (workload, work-life predictability, wellbeing, recognition, leadership support, and so on) for the Section~5 co-occurrence analysis. The codebook and every coded row carry full provenance, so coding is reproducible (Section~8).

\subsection{Construct coding pipeline}
We code constructs with an LLM (claude-opus-4-8, maximum effort), as augmentation rather than replacement of human coders (Ziems et al., 2024) and justified on cost and scale (Gilardi et al., 2023). Coding runs at two granularities: all 4,673 source units are coded at the unit level against the full codebook (the census behind the Section~5 distribution), while summaries are coded at the document level for which themes and voice types are present. For reliability we draw a gold sample (320 units), code it twice under independent prompts, and adjudicate disagreements into a resolved master against which production is checked. Chance-corrected agreement is high for the constructs our claims depend on (voice type κ = 0.96, sentiment α = 0.97, language κ = 0.98, explicit-suggestion κ = 0.91, theme α = 0.88). To test whether this reflects one model family's idiosyncrasies, a different family (GPT-4o) recodes the full corpus under the same codebook, and the core constructs hold cross-family (voice type κ = 0.82, sentiment α = 0.91, language κ = 0.80, explicit-suggestion κ = 0.87). Directness reproduces in neither check (within-family α = 0.37–0.48, cross-family α = 0.13). The prompts split the direct/hedged boundary differently, so we treat it as directional only (Section~5, Section~8.4). Every row is stamped with model, effort, prompt id, codebook version, run mode, and a fixed seed (20260613), so the pipeline is re-runnable (Section~8).

\subsection{Summary pipelines}
From the coded source we generate leader-facing summaries with the same model under three system prompts spanning a realistic deployment range. One production-grade model (claude-opus-4-8) fills all three roles, so the audit is a controlled simulation characterizing one current LLM; a GPT-4o replication of the terse config reproduces the same prevalence gradient (Section~8.4). A terse briefing (\texttt{exec\_terse}) returns at most three bullets for a twenty-second read; a balanced paragraph (\texttt{balanced}), three to five sentences; a mitigation prompt (\texttt{preserve}) tells the model to keep critical, dissenting, and minority concerns and to surface workload, work-life, protected-time, and wellbeing. We sample fifteen teams of moderate volume (22–42 source units each) so summarization entails real compression while coding stays tractable, yielding 426 source units and 45 summaries (15 × 3). Each summary is re-coded with the source's presence schema, so the two sides are comparable on the granularity they share.

\subsection{Audit metrics}
The audit asks whose contribution it keeps, the pivot from faithfulness (Maynez et al., 2020) to proportional representation (Dash et al., 2019). For each team \textit{theme retention} is the fraction of source themes reappearing in the summary, reported unweighted and salience-weighted by how many comments raise each theme; a theme's \textit{Representation Ratio} is its summary-presence rate over its source-presence rate, so a value below one marks under-representation. We also report Jaccard distance and Jensen-Shannon divergence between source and summary theme sets. To separate compression from muting, and because critical themes are raised in more comments, we fit a cluster-robust logit of theme survival on a critical-theme indicator, log source-unit count, and config, clustered by team, and benchmark critical-theme retention against two permutation nulls (salience-blind and salience-proportional). With only 15 clusters, below the conventional 30–40, we read the logit's p-values as approximate and lean on the permutation tests. The gap between salience-weighted and unweighted retention is the muting-beyond-compression signal: if a config sheds low-frequency themes faster, salience-weighted retention rises above the plain count, whereas mere shortening would move neither (Section~6).

\section{Results I: The Voice (RQ1)}

Among the 2,586 responses that contain free text, the two fields are filled at strikingly different rates: IMPROVE in 99.8\% (2,580 of 2,586) but WENT\_WELL only 80.9\% (2,093). Of rows with exactly one field filled, 493 give criticism without praise and just six the reverse, an asymmetry of roughly 82 to 1. Length agrees: IMPROVE answers average 108 characters against 70 for WENT\_WELL, and where both are filled the IMPROVE answer is longer 63\% of the time. Employees far more often leave the \textit{appreciation} field blank (Figure~1), inverting the silence literature's focus on problem-silence (Van Dyne et al., 2003; Sherf et al., 2021). We therefore read the asymmetry as how engaged commenters use this channel, not a universal claim about withholding praise.

\begin{figure*}[t]
  \centering
  \includegraphics[width=\textwidth]{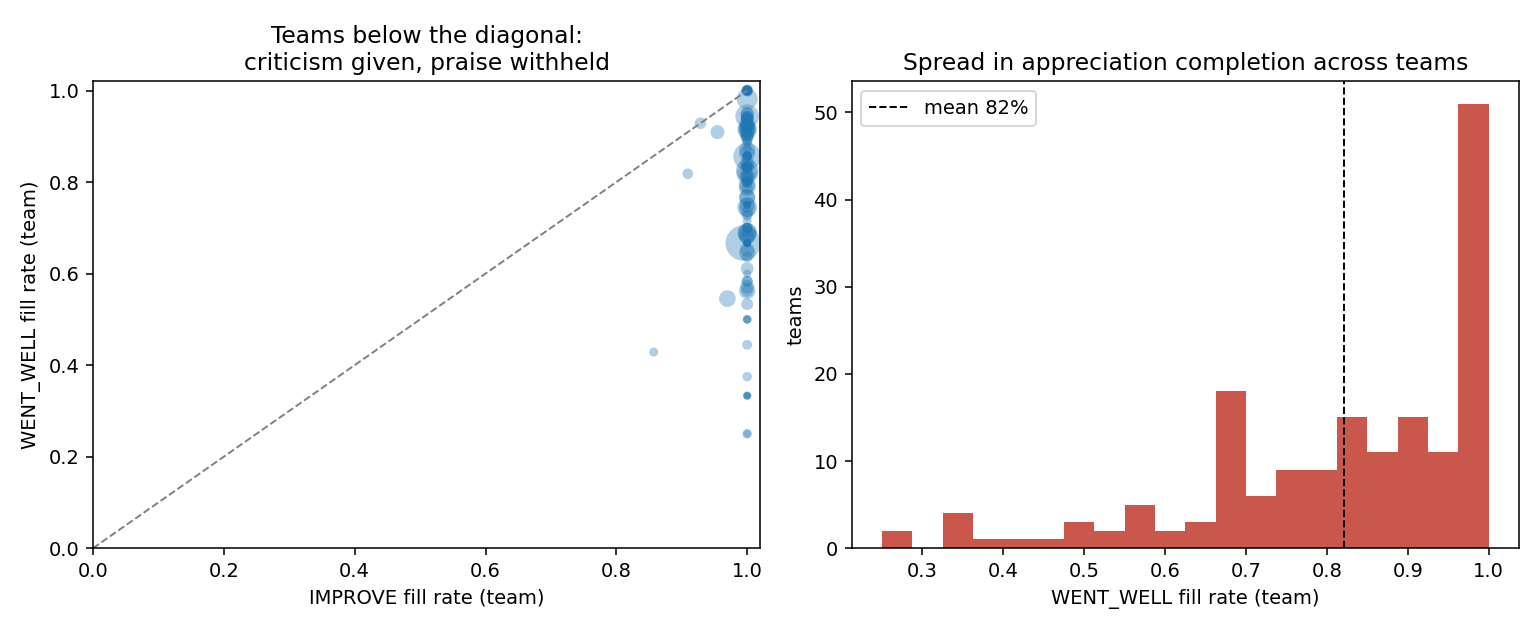}
  \caption{Appreciation asymmetry. Left: per-team IMPROVE vs WENT\_WELL completion, nearly all below the diagonal. Right: spread of team WENT\_WELL completion.}
  \label{fig:asymmetry}
\end{figure*}

Skipping praise also varies across teams. Among the 169 teams with at least three responses, WENT\_WELL completion averages 82\% but ranges from 25\% to 100\%, with 19 teams below 60\%, and 6 to 14\% of the variance sits at the team level (ICC 0.06–0.07; logistic latent-scale 0.14). Withholding appreciation is therefore partly a team trait, consistent with a psychological-safety reading, though the 82-fold asymmetry holds regardless of the clustering estimate.

Construct coding of the full corpus (4,673 units, Table~1) shows the IMPROVE channel is dominated by suggestions, not complaints: promotive voice outnumbers prohibitive 3.23 to 1, and open problem-statement is the rarest substantive register at about 12\% of units. Employees hedge problems and suggestions about equally, so hedging runs across the change-oriented channel rather than marking criticism alone, consistent with the self-protective indirectness of implicit voice theory (Detert \& Edmondson, 2011). Voice points sideways or at the system rather than up: improvement asks most often target processes and systems (38\%) and peers (27\%), while only 8\% reach senior managers. Praise, when given, names the whole team far more than any individual (1,094 went-well units against roughly 190 naming a person). Two minority signals matter later: about 13\% of units invoke a teaming norm such as protected time, and under 4\% carry a futility marker (“again”, “still”, “wie immer”), the trace of a point already raised and unaddressed. German improvement voice is coded as less direct than English at the source, a gap holding across passes though our least reproducible code, so we report it directionally. The channel is constructive but indirect, and its highest-stakes content is also its thinnest, the content an upward summary is most likely to compress.

\begin{table}[tb]
  \centering
  \caption{Voice-type distribution (4,673 coded units); in IMPROVE, promotive outnumbers prohibitive 3.23:1.}
  \label{tab:voicedist}
  \begin{tabular}{lrr}
    \toprule
    Voice type & n & Share \\
    \midrule
    Appreciative & 2,072 & 44.3\% \\
    Promotive & 1,897 & 40.6\% \\
    Prohibitive & 585 & 12.5\% \\
    Descriptive-neutral & 67 & 1.4\% \\
    Perfunctory & 52 & 1.1\% \\
    \bottomrule
  \end{tabular}
\end{table}

Because the project codes encode weekly waves, we can ask whether voice erodes as a project grinds on. Across the 44 projects with three or more waves (966 responses), wave rank shows no measurable trend in appreciation completion (odds ratio 1.05, p = 0.49), futility markers, suggestion rate, or response length (all p $>$ 0.1), and theme recurrence sits at chance. The asymmetry is present from the first wave and holds throughout, a standing feature rather than late-project fatigue.

Improvement comments are not a flat list of grievances; they cluster (Figure~2). The largest coupling in the corpus is workload with work-life predictability (lift 1.61, n = 99, p = 2.6×10⁻⁸): when employees raise overload, 29\% of those comments also raise predictability of time, and overload spills into wellbeing (lift 1.66, p = 0.004), echoing the overwork-and-predictability coupling Perlow (2012) documents. Two satellites orbit the workload hub: a recognition cluster, where requests for appreciation attach to leadership (lift 2.18, p = 0.0003) and communication, and a logistics cluster binding colocation to remote flexibility (lift 6.31). The linear chain from client pressure through workload to lost work-life that we expected was not supported (mean edge-lift 0.95), so we drop it. The structure is a workload-centred star, and that hub is exactly the content an upward summary must not lose. Whether it does is the question of Section~6.

\begin{figure*}[t]
  \centering
  \includegraphics[width=\textwidth]{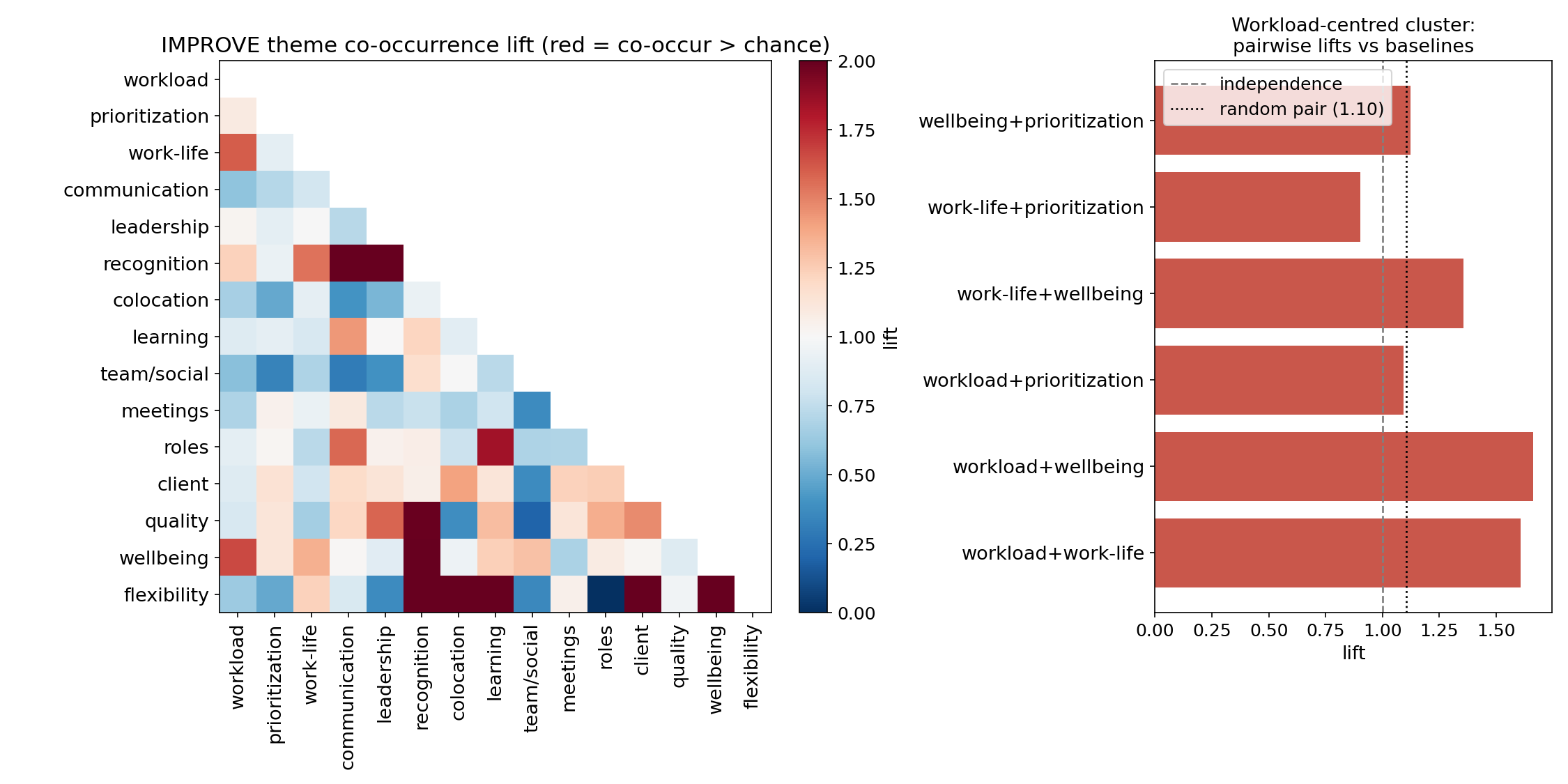}
  \caption{Improvement-voice architecture. Left: theme co-occurrence lift (red = above-chance pairs), workload/work-life/wellbeing at the hub. Right: the workload complex vs baselines.}
  \label{fig:architecture}
\end{figure*}

\section{Results II: Auditing What Survives the Summary (RQ2)}

The intuitive failure mode does not occur. Appreciative, promotive, and prohibitive voice are each present in all 15 summaries under every config, the three-bullet \texttt{exec\_terse} included. Even the tersest briefings name problems: one team is told to “stop teaming sessions that don't feel psychologically safe”, another to address “heavy reporting effort that ties up resources”. At the level of presence, the summaries are not whitewashed into praise, though presence is all this document-level coding certifies. It registers that a problem survived, not whether its intensity did, so a summary could retain a criticism while softening its tone, a flattening we cannot detect here (Section~8.4). If criticism is not what gets dropped, the loss must be measured one level down, at the theme level, the granularity the source and summary codings share.

Table~2 reports theme retention by config. The three-bullet \texttt{exec\_terse} keeps 61\% of a team's source themes; the paragraph configs keep about 71\% (paired Wilcoxon vs exec\_terse: balanced p = 0.0014, preserve p = 0.0058). Shorter summaries drop more. Salience-weighted retention, weighting each theme by how many comments raise it, exceeds the plain count in every config (0.77 vs 0.61 for exec\_terse), so the loss concentrates at the low-frequency end rather than spreading evenly. Distributional checks agree: \texttt{balanced} is most faithful to the source theme set (Jaccard 0.30), \texttt{exec\_terse} least (0.40).

\begin{table}[tb]
  \centering
  \caption{Theme retention by config (15 teams), with 95\% CIs; salience-weighted retention exceeds the plain count everywhere.}
  \label{tab:retention}
  \setlength{\tabcolsep}{4pt}
  \small
  \begin{tabular}{lcc}
    \toprule
    Config & Retention [95\% CI] & \shortstack{Salience-\\weighted} \\
    \midrule
    exec\_terse & 0.61 [0.55, 0.67] & 0.77 \\
    balanced & 0.71 [0.66, 0.76] & 0.87 \\
    preserve & 0.70 [0.65, 0.75] & 0.83 \\
    \bottomrule
  \end{tabular}
\end{table}

The next question is whether the dropped themes are the critical ones. The raw pattern says no, surprisingly: critical themes (carrying at least one prohibitive or negative-sentiment unit) are retained more than positive-only themes in every config (0.73 vs 0.49 under exec\_terse, Fisher p = 0.001). Read literally, the summarizer privileges criticism, but this is a confound. Critical themes are raised in more comments (6.62 source units on average against 4.20 for positive-only), and singletons make up 6\% of critical themes against 24\% of positive ones. Problems get repeated across a team; praise is more often a one-off. Once we control for how many comments raise a theme, the valence effect disappears: in a cluster-robust logit of theme survival (n = 564 team-by-theme-by-config cells, clustered by team), log source-unit count carries +1.61 (p $<$ 0.001) while the critical indicator sits at +0.22 (p = 0.55), indistinguishable from zero (Table~3). A permutation test agrees: criticism is not suppressed beyond what compression alone produces. Because the 15 clusters underpower the logit, the permutation result is the primary basis for this null and the logit corroborates it. The summarizer is not a sentiment filter but a popularity filter.

\begin{table}[tb]
  \centering
  \caption{Cluster-robust logit of theme survival (564 cells, clustered by team). Frequency dominates; valence has no independent effect.}
  \label{tab:logit}
  \begin{tabular}{lcc}
    \toprule
    Predictor & Coefficient & p \\
    \midrule
    log(source units) & +1.61 & $<$ 0.001 \\
    Critical theme & +0.22 & 0.55 \\
    \bottomrule
  \end{tabular}
\end{table}

If survival tracks prevalence, the voices that lose are those carried by few, short, or non-majority-language comments. All three gaps are visible (Figure~3). On frequency, a theme raised in three or more comments is retained 0.74 [0.66, 0.80] under exec\_terse while a single-comment theme is retained 0.14 [0.05, 0.31], non-overlapping intervals and a 5.3-fold gap, so a singleton is dropped about 86\% of the time. On length, themes raised only in short comments (at or below the 65-character median) are retained 0.27 against 0.65 for themes with at least one elaborated comment, a 2.4-fold gap. On language, German-only themes survive at 0.20 against 0.61 overall, though all five cells are singletons and the result is directional. The same ranking appears theme by theme: the most under-represented are learning and development (representation ratio 0.29), role clarity (0.31), recognition (0.33), and leadership support (0.36), all lower-frequency satellites, while the high-frequency workload hub passes through nearly intact (1.07).

\begin{figure*}[tp]
  \centering
  \includegraphics[width=\textwidth]{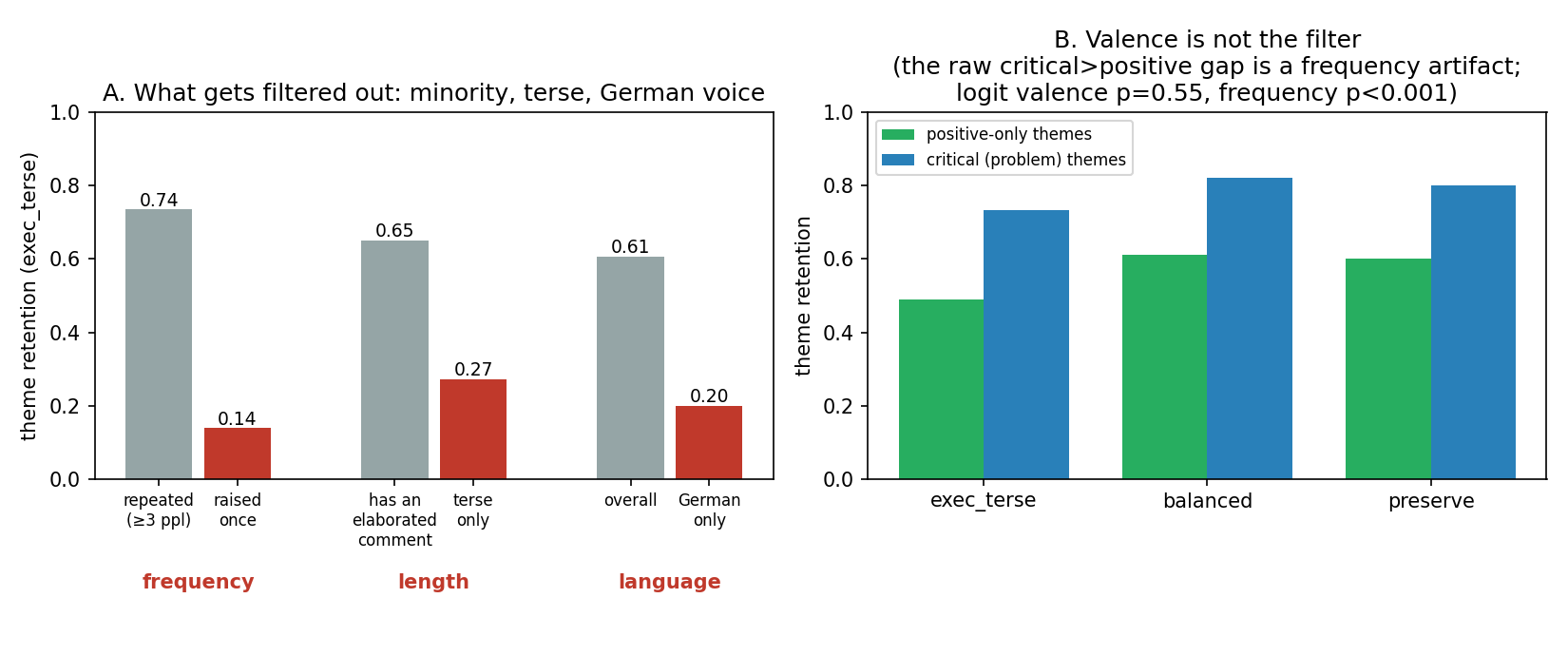}
  \caption{The popularity filter. Panel A: retention by frequency, length, and language. Panel B: critical themes survive as often as positive ones (prevalence artifact; valence p = 0.55).}
  \label{fig:filter}
\end{figure*}

The harm is a prevalence-driven bias against concerns held by few: a balanced-looking summary gives a leader no cue that a lone wellbeing flag or single German comment was~ever~there.

\section{Results III: Mitigation and Design Principles (RQ2)}

The \texttt{preserve} config tests whether instruction can repair the filter. It names workload, work-life, and wellbeing and tells the model to keep critical, dissenting, and minority concerns. On the themes it names, it works: retention of the protected set rises from 0.725 (exec\_terse) to 0.850 (balanced) to 0.900 (preserve) (paired Wilcoxon preserve vs exec\_terse p = 0.015), and German-only retention doubles from 0.20 to 0.40. But the gain is precise to the instruction: \texttt{preserve} does not beat \texttt{balanced} across the full set of critical themes (0.80 vs 0.82), and does nothing for the frequency and length axes it never mentions, where a singleton outside the protected set drops at about the same rate as before. A prompt-level fix is only as good as its named targets; it cannot repair a structural bias it does not enumerate, so the popularity filter survives any fix that protects themes by name.

Five design principles follow, each tied to a result. (1) \textit{Frequency is not importance.} Survival is driven by repetition, not content (log-units p $<$ 0.001), so a concern raised once is dropped about 86\% of the time; a pipeline must not read “said by many” as “matters most.” (2) \textit{Protect the singleton.} Instruct summarizers to keep single-comment concerns, and audit singleton retention as a first-class metric. (3) \textit{Preserve terse voice.} Short comments drop about 2.4 times more often than elaborated ones, so terse units should be up-weighted before summarizing. (4) \textit{Language parity.} German-only content survives at roughly a third of the corpus rate; non-majority languages need explicit parity checks. (5) \textit{Audit on the right axes.} A sentiment-only audit gives this pipeline a clean bill of health while it filters out minority, terse, and German voice, so audits must measure retention by frequency, length, and language, rather than tone alone. We instantiate transparency as a \textit{voice-retention card}: a short, disaggregated report of retention by construct, language, and length (and, where the data permit, by social group and role), following model-card (Mitchell et al., 2019) and datasheet (Gebru et al., 2021) templates, and consistent with treating the tool as a “fallible companion technology” (Gal et al., 2020).

\section{Discussion}

\subsection{Theoretical implications}
Our findings extend voice/silence theory to an algorithmic setting. The literature models silence as a human, climate-level phenomenon (Morrison \& Milliken, 2000; Detert \& Edmondson, 2011); we identify a \textit{new locus}, the summarization layer, where voice that was successfully raised is re-suppressed before reaching a decision-maker. The axis of re-suppression is one voice theory would not predict: \textit{frequency} rather than \textit{valence}, adding a \textit{prevalence} dimension to a literature framed by motive (Van Dyne et al., 2003) and climate (Morrison \& Milliken, 2000). This positions the LLM as a consequential intermediary in the upward flow of voice (Detert et al., 2013; Burris et al., 2025) and operationalizes the promotive/prohibitive distinction (Liang et al., 2012) on naturalistic text rather than survey self-report. The two findings also compound: employees already under-supply praise and soften much of their voice (Section~5), so the voice entering the summarizer is thin where it is most fragile, and the popularity filter then removes the thinnest signals first, leaving the leader a doubly majority-biased picture.

\subsection{Responsible-AI implications}
Reframing summarization failure from \textit{factual error} (Maynez et al., 2020) to \textit{inequitable retention by prevalence} makes representational harm (Blodgett et al., 2020) measurable in an organizational pipeline, complementing fairness-in-summarization work confined to social-media and demographic settings (Dash et al., 2019; Keswani \& Celis, 2021). The harm we measure is not the muting of criticism responsible-AI intuition anticipates; critical content largely survives. What the pipeline sheds is infrequent voice, a subtler failure because the summary reads as balanced and complete. Because automation bias (Parasuraman \& Manzey, 2010) makes the omission invisible, this minority-voice gap can become a structural decision blind spot. Prevalence also intersects the social-group axes a justice lens names. Language is one such axis (Joshi et al., 2020), and here German-only voice is attenuated along the same prevalence gradient, so a prevalence-blind summarizer can turn a linguistic minority into an unheard one. Where employee demographics are available, the same Representation Ratio can be computed per group, so demographic parity is a direct extension of our metric rather than a separate method, an audit our corpus cannot run but the card is built to hold.

\subsection{Governance and surveillance}
Beyond representational harm, the pipeline inherits the people-analytics perils an opaque, dissent-dropping summary would aggravate (Tursunbayeva et al., 2018; Giermindl et al., 2022; Bernstein, 2012; Zuboff, 2015). Disclosing the voice-retention card to employees, not only to leaders, reframes the pipeline from one-directional surveillance (Zuboff, 2015) toward mutual accountability: the metric that warns leaders what a summary dropped also tells employees whether their concern reached the top. This is a direct turn on Bernstein's (2012) transparency paradox, since the card discloses exactly what the summary hides. Our auditable retention metric and voice-retention card move that critique from diagnosis toward design (Gal et al., 2020).

\subsection{Limitations}
Several bounds apply. The evidence is one corpus from one firm, and only its self-selected free-text responses, so the asymmetry and filter are strong descriptive facts about employees who chose to comment here. Teams are identified by base project code, and our frequency measure counts comments raising a theme, not distinct individuals; because a project code's volume also reflects how many waves it ran, we decompose it into within-wave breadth and across-wave recurrence and find both predict survival about equally (each p $<$ 0.001; equal-weight Wald p = 0.85), so the filter tracks mention volume however it accumulates. The temporal axis is wave rank, not calendar time, so within-person decay is out of reach (Section~5). Construct codes come from an LLM, bounded by a doubly coded, adjudicated gold sample (core-construct κ and α between 0.88 and 0.98; directness, the one weak field at α = 0.48, is directional only). Because the same model family generates and codes the summaries, one might suspect circularity. For coding, a second model family reproduces the core constructs across the full corpus (Section~4.2), so the codes are not an artifact of one family. For generation, any such bias runs against our finding, since a coder primed to recognize its own model's concepts would over-credit matches and inflate retention, whereas we report a deficit. More basically, a theme absent from the summary text cannot be coded present, so the drop occurs at generation, is directly observable, and is independent of coding; circularity could dampen the measured gap but cannot manufacture the singleton-attrition pattern. The language and length axes are directional: the German-only result rests on five cells and, because summaries are generated in English by design, concerns whether German-origin content survives, not the language a leader reads. Re-running the terse config with a different summarizer (GPT-4o) reproduces the prevalence gradient (singleton 0.14 vs common 0.57; log-units p $<$ 0.001), so the popularity filter is not specific to one model; broader cross-model audits remain future work. Finally, the claims are descriptive: with no outcome or demographic data we make no causal claims, and retention is measured on theme presence, a meaningful floor, since a theme absent from the summary has surely lost its nuance too, so our attrition estimate is, if anything, conservative. By the same token, presence cannot register the \textit{tone} of what survives: a retained criticism may still be softened, so our “criticism survives” claim concerns presence, not preserved intensity, and detecting such tone-flattening is future work.

\section{Conclusion}

We asked whose voice survives the summary. Three answers emerged. Employees in this corpus withhold praise far more than criticism, an asymmetry of about 82 to 1 that inverts the usual expectation about where silence falls. When an LLM summarizes that voice upward, it filters by popularity: criticism survives almost everywhere, while concerns raised by only one or two people are dropped, terse and German-only voice (the latter directional) with them. A targeted prompt recovers the themes it is told to protect but leaves the frequency and length bias in place, so prompt-level fixes are partial by construction.

The contribution is a way to see this and a starting point for fixing it: a Voice Retention and Representation Ratio metric that measures equity of retention by prevalence, length, and language; field evidence that the harm is prevalence-driven, not the valence bias intuition expects; and design guidance, including a disaggregated voice-retention card. As employee listening becomes LLM-mediated by default, the summarization layer decides which concerns a leader ever sees. A pipeline that equates frequency with importance will keep telling leaders what is already said most often, and drop the lone voice that decades of voice research suggest an organization can least afford to lose.

\section*{References}

{\small
\setlength{\parindent}{0pt}
\refitem{Bernstein, E. S. (2012). The transparency paradox: A role for privacy in organizational learning and operational control. \textit{Administrative Science Quarterly, 57}(2), 181–216.}
\refitem{Blodgett, S. L., Barocas, S., Daumé III, H., \& Wallach, H. (2020). Language (technology) is power: A critical survey of “bias” in NLP. In \textit{Proceedings of the 58th Annual Meeting of the Association for Computational Linguistics} (pp. 5454–5476).}
\refitem{Burris, E. R. (2012). The risks and rewards of speaking up: Managerial responses to employee voice. \textit{Academy of Management Journal, 55}(4), 851–875.}
\refitem{Burris, E. R., Howell, T. M., Stubben, S. R., \& Welch, K. T. (2025). Not just hearsay and rumor: How managers (actually) perceive the credibility of secondhand accounts of employee voice. \textit{Academy of Management Journal.} Advance online publication.}
\refitem{Dash, A., Shandilya, A., Biswas, A., Ghosh, K., Ghosh, S., \& Chakraborty, A. (2019). Summarizing user-generated textual content: Motivation and methods for fairness in algorithmic summaries. \textit{Proceedings of the ACM on Human-Computer Interaction, 3}(CSCW), Article 172.}
\refitem{Detert, J. R., Burris, E. R., Harrison, D. A., \& Martin, S. R. (2013). Voice flows to and around leaders: Understanding when units are helped or hurt by employee voice. \textit{Administrative Science Quarterly, 58}(4), 624–668.}
\refitem{Detert, J. R., \& Edmondson, A. C. (2011). Implicit voice theories: Taken-for-granted rules of self-censorship at work. \textit{Academy of Management Journal, 54}(3), 461–488.}
\refitem{Edmondson, A. C. (1999). Psychological safety and learning behavior in work teams. \textit{Administrative Science Quarterly, 44}(2), 350–383.}
\refitem{Edmondson, A. C., \& Bransby, D. P. (2023). Psychological safety comes of age: Observed themes in an established literature. \textit{Annual Review of Organizational Psychology and Organizational Behavior, 10}, 55–78.}
\refitem{Gal, U., Jensen, T. B., \& Stein, M.-K. (2020). Breaking the vicious cycle of algorithmic management: A virtue ethics approach to people analytics. \textit{Information and Organization, 30}(2), Article 100301.}
\refitem{Gebru, T., Morgenstern, J., Vecchione, B., Vaughan, J. W., Wallach, H., Daumé III, H., \& Crawford, K. (2021). Datasheets for datasets. \textit{Communications of the ACM, 64}(12), 86–92.}
\refitem{Giermindl, L. M., Strich, F., Christ, O., Leicht-Deobald, U., \& Redzepi, A. (2022). The dark sides of people analytics: Reviewing the perils for organisations and employees. \textit{European Journal of Information Systems, 31}(3), 410–435.}
\refitem{Gilardi, F., Alizadeh, M., \& Kubli, M. (2023). ChatGPT outperforms crowd workers for text-annotation tasks. \textit{Proceedings of the National Academy of Sciences, 120}(30), e2305016120.}
\refitem{Joshi, P., Santy, S., Budhiraja, A., Bali, K., \& Choudhury, M. (2020). The state and fate of linguistic diversity and inclusion in the NLP world. In \textit{Proceedings of the 58th Annual Meeting of the Association for Computational Linguistics} (pp. 6282–6293).}
\refitem{Keswani, V., \& Celis, L. E. (2021). Dialect diversity in text summarization on Twitter. In \textit{Proceedings of the Web Conference 2021} (pp. 3802–3814).}
\refitem{Liang, J., Farh, C. I. C., \& Farh, J.-L. (2012). Psychological antecedents of promotive and prohibitive voice: A two-wave examination. \textit{Academy of Management Journal, 55}(1), 71–92.}
\refitem{Maynez, J., Narayan, S., Bohnet, B., \& McDonald, R. (2020). On faithfulness and factuality in abstractive summarization. In \textit{Proceedings of the 58th Annual Meeting of the Association for Computational Linguistics} (pp. 1906–1919).}
\refitem{Mitchell, M., Wu, S., Zaldivar, A., Barnes, P., Vasserman, L., Hutchinson, B., Spitzer, E., Raji, I. D., \& Gebru, T. (2019). Model cards for model reporting. In \textit{Proceedings of the Conference on Fairness, Accountability, and Transparency} (pp. 220–229).}
\refitem{Morrison, E. W., \& Milliken, F. J. (2000). Organizational silence: A barrier to change and development in a pluralistic world. \textit{Academy of Management Review, 25}(4), 706–725.}
\refitem{Parasuraman, R., \& Manzey, D. H. (2010). Complacency and bias in human use of automation: An attentional integration. \textit{Human Factors, 52}(3), 381–410.}
\refitem{Perlow, L. A. (2012). \textit{Sleeping with your smartphone: How to break the 24/7 habit and change the way you work.} Harvard Business Review Press.}
\refitem{Perlow, L. A., \& Williams, S. (2003). Is silence killing your company? \textit{Harvard Business Review, 81}(5), 52–58, 128.}
\refitem{Sherf, E. N., Parke, M. R., \& Isaakyan, S. (2021). Distinguishing voice and silence at work: Unique relationships with perceived impact, psychological safety, and burnout. \textit{Academy of Management Journal, 64}(1), 114–148.}
\refitem{Tamme, T., Hantel, A., Saatkamp, M., \& Scheuer, A. P. (2026). AI-native qualitative interviews at scale: Leveraging conversational agents for organisational research. \textit{ECIS 2026 TREOs}, 20.}
\refitem{Tursunbayeva, A., Di Lauro, S., \& Pagliari, C. (2018). People analytics—A scoping review of conceptual boundaries and value propositions. \textit{International Journal of Information Management, 43}, 224–247.}
\refitem{Van Dyne, L., Ang, S., \& Botero, I. C. (2003). Conceptualizing employee silence and employee voice as multidimensional constructs. \textit{Journal of Management Studies, 40}(6), 1359–1392.}
\refitem{Zhang, T., Ladhak, F., Durmus, E., Liang, P., McKeown, K., \& Hashimoto, T. B. (2024). Benchmarking large language models for news summarization. \textit{Transactions of the Association for Computational Linguistics, 12}, 39–57.}
\refitem{Ziems, C., Held, W., Shaikh, O., Chen, J., Zhang, Z., \& Yang, D. (2024). Can large language models transform computational social science? \textit{Computational Linguistics, 50}(1), 237–291.}
\refitem{Zuboff, S. (2015). Big other: Surveillance capitalism and the prospects of an information civilization. \textit{Journal of Information Technology, 30}(1), 75–89.}
}

\end{document}